\documentclass{article}
\usepackage{spconf,amsmath,graphicx}
\usepackage{booktabs}
\usepackage{multirow}
\usepackage{pgfplots, pgfplotstable}
\usepackage{braket}
\usepackage{makecell}
\usepackage[inline]{enumitem}
\usepackage{cite}
\pgfplotsset{compat=1.14}

\def\D{{\mathcal{D}}}
\DeclareMathOperator*{\argmax}{argmax}

\title{Self-Training for End-to-End Speech Recognition}
\name{Jacob Kahn, Ann Lee, Awni Hannun}
\address{Facebook AI Research}
\begin{document}
%
\maketitle
\begin{abstract}
We revisit self-training in the context of end-to-end speech recognition. We demonstrate that training with pseudo-labels can substantially improve the accuracy of a baseline model. Key to our approach are a strong baseline acoustic and language model used to generate the pseudo-labels, filtering mechanisms tailored to common errors from sequence-to-sequence models, and a novel ensemble approach to increase pseudo-label diversity. Experiments on the LibriSpeech corpus show that with an ensemble of four models and label filtering, self-training yields a 33.9\% relative improvement in WER compared with a baseline trained on 100 hours of labelled data in the noisy speech setting. In the clean speech setting, self-training recovers 59.3\% of the gap between the baseline and an oracle model, which is at least 93.8\% relatively higher than what previous approaches can achieve. 
\end{abstract}

\vspace{-3mm}
\section{Introduction}
\label{sec:intro}


Building automatic speech recognition (ASR) systems requires a large amount of transcribed data. Compared with hybrid models, the performance of end-to-end models significantly degrades as the amount of available training data decreases~\cite{luscher2019rwth}. Transcribing large quantities of audio is both expensive and time-consuming, and thus many semi-supervised training approaches have been proposed to take advantage of abundant unpaired audio and text data. One such approach, self-training~\cite{scudder1965probability}, uses noisy labels generated from a model trained on a much smaller labelled data set.

We revisit self-training in the context of sequence-to-sequence models. Self-training has not been carefully studied in end-to-end speech recognition. We start from training a strong baseline acoustic model on a small paired data set and performing stable decoding~\cite{hannun2019sequence} with a language model (LM) trained on a large-scale text corpus to generate pseudo-labels. We evaluate one heuristic and one confidence-based method for pseudo-label filtering~\cite{charlet2001confidence,wessel2004unsupervised,vesely2013semi,vesely2017semi} tailored to the mistakes often encountered with sequence-to-sequence models. In addition, we propose an ensemble approach that combines multiple models during training to improve label diversity and keep the model from being overly confident to noisy pseudo-labels.

We demonstrate the effectiveness of self-training on LibriSpeech~\cite{panayotov2015librispeech}, a publicly available corpus of read speech. In particular, we study the trade-off between the amount of unpaired audio data, the quality of the pseudo-labels, and the model performance. We find that in the clean speech setting, as the label quality is high, the model performance depends heavily on the amount of data. In the noisy speech setting, a proper filtering mechanism is essential for removing noisy pseudo-labels. In addition, using an ensemble of models can be complementary to filtering.

Compared with other semi-supervised methods with sequence-to-sequence models~\cite{hori2019cycle,baskar2019self}, we show that self-training achieves a 93.8\% relatively higher WER recovery rate (WRR)~\cite{ma2008unsupervised} on the clean test set, a metric indicating how much the gap between a supervised baseline and an oracle can be bridged. One goal of this work is to provide a publicly-available and reproducible benchmark to which future semi-supervised approaches in ASR can compare.

\section{Model}

Our sequence-to-sequence model is an encoder-decoder architecture with attention~\cite{bahdanau2014neural, cho2014learning}. Let $X = [X_1, \ldots, X_T]$ be the frames of speech with transcription $Y = [y_1, \ldots, y_U]$. The encoder maps $X$ into a key-value hidden representation:
\begin{equation}
    \begin{bmatrix} K\\ V \end{bmatrix} = \text{encode}(X)
\end{equation}
where $K = [K_1, \ldots, K_T]$ are the keys and $V = [V_1 \ldots, V_T]$ are the values. We use a fully convolutional encoder with time-depth separable (TDS) blocks proposed in~\cite{hannun2019sequence}. The decoder is given by
\begin{align}
    Q_u &= RNN(y_{u-1}, Q_{u-1}) \label{eq:attend} \\
    S_u &= \text{attend}(Q_{u}, K, V) \\
    P(y_u \mid X, y_{< u}) &= h(S_u, Q_u).
\end{align}
The RNN encodes the previous token and query vector $Q_{u-1}$ to produce the next query vector. The attention mechanism produces a summary vector $S_u$ with a simple inner product:
\begin{align}
    \text{attend}(K, V, Q) =  V \cdot \text{softmax}\left(\frac{1}{\sqrt{d}} K^\top Q \right)
\end{align}
where $d$ is the hidden dimension of $K$ (as well as $Q$ and $V$). $h(\cdot)$ computes a distribution over the output tokens.

\subsection{Inference}
\label{sec:inference}

During inference, we carry out beam search to search for the most likely hypothesis according to the sequence-to-sequence model ($P_{\textrm{AM}}$) and an external language model ($P_{\textrm{LM}}$):
\begin{equation}
\label{eq:inference}
    \bar{Y} = \argmax_Y \log P_{\textrm{AM}}(Y \mid X) + \alpha \log P_{\textrm{LM}}(Y) + \beta |Y|
\end{equation}
where $\alpha$ is the LM weight, and $\beta$ is a token insertion term for avoiding the early stopping problem common for sequence-to-sequence models~\cite{chorowski2016towards}. We follow the techniques in~\cite{hannun2019sequence} to improve the efficiency and stability of the decoder. One such technique is to only propose end-of-sentence (\textsc{EOS}) when the corresponding probability satisfies
\begin{align}
\label{eq:eos}
    \log P_u(\textsc{EOS} \mid y_{<u}) > \gamma \cdot \max_{c \ne \textsc{EOS}} \log P_u(c \mid y_{<u})
\end{align}
where $\gamma$ is a hyper-parameter that can be tuned.

\vspace{-1.5mm}
\section{Semi-supervised Self-training}
\vspace{-1mm}
In a supervised learning setting, we have access to a paired data set $\D = \{(X_1, Y_1), \ldots, (X_n, Y_n)\}$. We train a model on $\D$ by maximizing the likelihood of the ground-truth transcriptions given their corresponding utterances:
\begin{equation}
\label{eq:supobj}
     \sum_{(X, Y) \in \D}  \log P(Y \mid X).
\end{equation}
In a semi-supervised setting, we have an unlabelled audio data set $\mathcal{X}$ and an unpaired text data set $\mathcal{Y}$ in addition to $\D$. We first train an acoustic model on $\D$ by maximizing the objective in Equation~\ref{eq:supobj}. We also train an LM on $\mathcal{Y}$. We then combine the two models to generate a pseudo-label for each unlabelled example by solving Equation~\ref{eq:inference} and obtain a pseudo paired data set $\bar{\D} = \{(X_i, \bar{Y}_i) \mid X_i \in \mathcal{X}\}$. A new acoustic model can be trained on the combination of $\D$ and $\bar{\D}$ with the objective
\begin{equation}
\label{eq:selftrainobj}
     \sum_{(X, Y) \in \D}  \log P(Y \mid X) + \sum_{(X, \bar{Y}) \in \bar{\D}}  \log P(\bar{Y} \mid X).
\end{equation}

\vspace{-4mm}
\subsection{Filtering}
\label{sec:filter}

The pseudo-labelled data set $\bar{\D}$ contains noisy transcriptions. Filtering is a commonly used technique to achieve the right balance between the size of $\bar{\D}$ and the noise in the pseudo-labels. We design two heuristic-based filtering functions specific to sequence-to-sequence models, which can be further combined with conventional confidence-based filtering, and apply both filtering techniques on the sentence level.

Sequence-to-sequence models are known to easily fail at inference in two ways: looping and early stopping~\cite{chorowski2016towards}. We filter for the looping by removing pseudo-labels which contain an $n$-gram repeated more than $c$ times. As described in Section~\ref{sec:inference}, we deal with early stopping by only keeping hypotheses with an \textsc{EOS} probability above a threshold. However, we filter examples where the beam search terminates without finding any complete hypotheses.

Additionally, for each pseudo-label, we compute the length-normalized log likelihood from the sequence-to-sequence model as the confidence score:
\begin{equation*}
     \text{ConfidenceScore}(\bar{Y}_i) = \frac{\log P_{\textrm{AM}}(\bar{Y}_i \mid X_i)}{|\bar{Y}_i|}
\end{equation*}
where $|\bar{Y}_i|$ is the number of tokens in the utterance.

\vspace{-1mm}
\subsection{Ensembles}
\label{sec:ensembles}

Model combination often helps reduce word error rates in ASR. One way to utilize multiple models in self-training is to combine the model scores during inference to generate a single pseudo-labelled set with higher quality.
However, as $M$ increases, the decoding process becomes heavyweight.

Instead, we propose \textit{sample ensemble}. Given $M$ bootstrapped acoustic models, we generate a pseudo-labelled data set, $\bar{\D}_m$, for each model in parallel. We then combine all $M$ sets of pseudo-labels with uniform weights and optimize the following objective during training
\begin{equation*}
     \sum_{(X, Y) \in \D}  \log P(Y \mid X) + \frac{1}{M} \sum_{m=1}^M \sum_{(X, \bar{Y}) \in \bar{\D}_m}  \log P(\bar{Y} \mid X) .
\end{equation*}

In the implementation, we first train $M$ models on $\D$ using different randomly initialized weights. We generate $\bar{\D}_m$ with hyper-parameters tuned with each model, respectively. During training, we uniformly sample a pseudo-label from one of the $M$ models as the target in every epoch.

\section{Experiments}
\subsection{Data}
\label{sec:data}

All experiments are performed on the LibriSpeech corpus~\cite{panayotov2015librispeech}. We use the ``train-clean-100" set containing 100 hours of clean speech as the paired data set. We perform experiments in two settings. In the clean speech setting, we use 360 hours of clean speech in the ``train-clean-360" set as the unpaired audio set, and in the noisy speech setting, we use 500 hours of noisy speech in the ``train-other-500" set. We report results on the standard dev and test clean/other (noisy) sets. 

The standard LM training text in LibriSpeech is derived from 14,476 public domain books~\cite{panayotov2015librispeech}. To make the learning problem more realistic for self-training, we remove all books related to the acoustic training data from the LM training data, resulting in a removal of 997 books. We apply sentence segmentation using the NLTK toolkit~\cite{loper2002nltk} and normalize the text by lower-casing, removing punctuation except for the apostrophe, and replacing hyphens with spaces. We do not replace non-standard words with a canonical verbalized form. We find that the resulting LMs achieve comparable perplexity to LMs trained on the standard corpus on the dev sets.

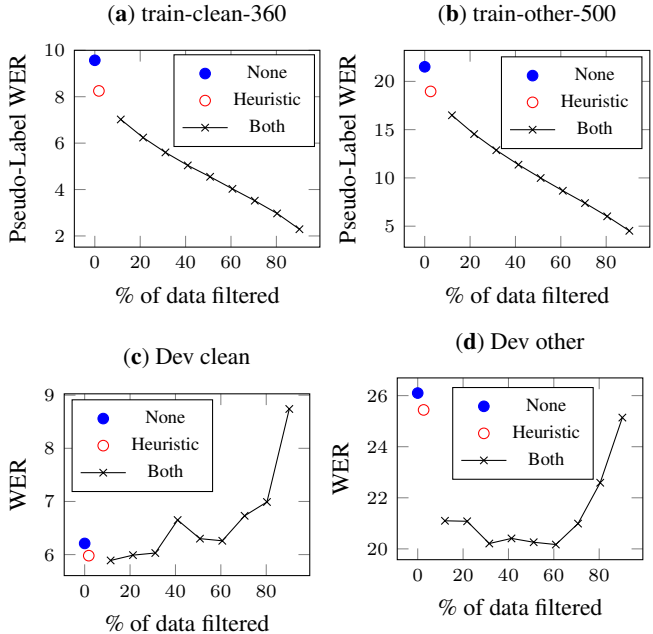
\begin{figure}
\begin{minipage}{.49\linewidth}
    \centering
  \begin{tikzpicture}
\begin{axis}[
  title={\small({\bf a}) train-clean-360},
  xlabel={\small \% of data filtered },
  ylabel={\small Pseudo-Label WER},
  ymax=10,
  xtick={0, 20, 40, 60, 80, 100},
  width=1.15\columnwidth,
  height=4.2cm,
  ticklabel style={font=\scriptsize},
  legend style={font=\scriptsize},
]
\addplot [only marks, color=blue] coordinates { (0.0,9.57) };
\addplot [only marks,mark=o, color=red] coordinates { (1.8,8.25) };
\addplot [mark=x] table [y=label, x=percentage]{filter_clean_v2.dat};

\addlegendentry{None}
\addlegendentry{Heuristic}
\addlegendentry{Both}

\end{axis}
\end{tikzpicture}
\end{minipage}
\begin{minipage}{.49\linewidth}
    \centering
  \begin{tikzpicture}
\begin{axis}[
  title={\small({\bf b}) train-other-500},
  xlabel={\small \% of data filtered },
  ylabel={\small Pseudo-Label WER},
  xtick={0, 20, 40, 60, 80, 100},
  width=1.15\columnwidth,
  height=4.2cm,
  ticklabel style={font=\scriptsize},
  legend style={font=\scriptsize},
  legend pos=north east
]
\addplot [only marks, color=blue] coordinates { (0.0,21.51) };
\addplot [only marks,mark=o, color=red] coordinates { (2.6,18.96) };
\addplot [mark=x] table [y=label, x=percentage]{filter_other_v2.dat};

\addlegendentry{None}
\addlegendentry{Heuristic}
\addlegendentry{Both}

\end{axis}
\end{tikzpicture}
\end{minipage}\vspace{-3mm}
\begin{minipage}{.49\linewidth}
    \centering
  \begin{tikzpicture}
\begin{axis}[
  title={\small({\bf c}) Dev clean},
  xlabel={\small \% of data filtered },
  ylabel={\small WER},
  xtick={0, 20, 40, 60, 80, 100},
  width=1.15\columnwidth,
  height=4cm,
  ticklabel style={font=\scriptsize},
  legend style={font=\scriptsize},
  legend pos=north west
]
\addplot [only marks,color=blue] coordinates { (0.0,6.21) };
\addplot [only marks,mark=o,color=red] coordinates { (1.8,5.98) };
\addplot [mark=x] table [y=devclean, x=percentage]{filter_clean_v2.dat};

\addlegendentry{None}
\addlegendentry{Heuristic}
\addlegendentry{Both}

\end{axis}
\end{tikzpicture}
\end{minipage}
\begin{minipage}{.49\linewidth}
    \centering
  \begin{tikzpicture}
\begin{axis}[
  title={\small({\bf d}) Dev other},
  xlabel={\small \% of data filtered },
  ylabel={\small WER},
  xtick={0, 20, 40, 60, 80, 100},
  width=1.15\columnwidth,
  height=4cm,
  ticklabel style={font=\scriptsize},
  legend style={font=\scriptsize,at={(0.8,0.95)}},
]
\addplot [only marks, color=blue] coordinates { (0.0,26.1) };
\addplot [only marks,mark=o, color=red] coordinates { (2.6,25.44) };
\addplot [mark=x] table [y=devother, x=percentage]{filter_other_v2.dat};

\addlegendentry{None}
\addlegendentry{Heuristic}
\addlegendentry{Both}

\end{axis}
\end{tikzpicture}
\end{minipage}
\vspace{-5mm} 
\caption{Results of different filtering functions and the corresponding pseudo-label quality ((\textbf{a}), (\textbf{b})) and model performance with LM beam search decoding ((\textbf{c}), (\textbf{d})) in clean ((\textbf{a}), (\textbf{c})) and noisy ((\textbf{b}), (\textbf{d})) settings, averaged across three runs. We vary the threshold on the confidence score to filter data at various deciles. (\textit{Both:} heuristic and confidence-based filters)}
\label{fig:filtering}
\vspace{-4mm}
\end{figure}


\subsection{Experimental Setting}

Our encoder consists of nine TDS blocks in groups of three, each with 10, 14 and 16 channels and a kernel width of 21. Other architectural details are the same as~\cite{hannun2019sequence}. We use the \emph{SentencePiece} toolkit~\cite{kudo2018sentencepiece} to compute 5,000 word pieces from the transcripts in ``train-clean-100" as the target tokens.


We follow the same training process as in~\cite{hannun2019sequence} with soft-window pre-training and teacher-forcing with 20\% dropout, 1\% random sampling, 10\% label smoothing and 1\% word piece sampling for regularization. We use a single GPU with a batch size of 16 when training baselines, and 8 GPUs when training with pseudo-labels. We use SGD without momentum for 200 epochs with a learning rate of 0.05, decayed by 0.5 every 40 epochs when using one GPU or 80 epochs for 8 GPUs. Experiments are done in the \emph{wav2letter++} framework~\cite{pratap2019wav2letter++}.

We train a word piece convolutional LM (ConvLM) using the same model architecture and training recipe as \cite{zeghidour2018fully}. All beam search hyper-parameters are tuned on the dev sets before generating the pseudo-labels. When training models with the combined paired and pseudo-labelled data sets, we start from random initialization instead of two-stage fine-tuning.

\vspace{-2mm}
\subsection{Results}

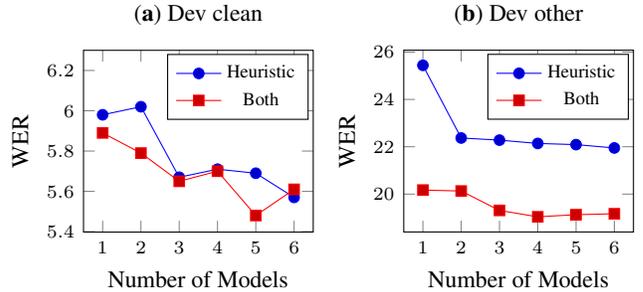
\begin{figure}
\begin{minipage}{.49\linewidth}
    \centering
  \begin{tikzpicture}
\begin{axis}[
  title={\small({\bf a}) Dev clean},
  xlabel={\small Number of Models},
  ylabel={\small WER},
  ymax=6.3,
  xtick=data,
  width=1.1\columnwidth,
  height=4cm,
  ticklabel style={font=\scriptsize},
  legend style={font=\scriptsize}
]
\addplot table [y=clean_heuristic, x=nmodel]{ensemble_lm.dat};
\addplot table [y=clean_filter, x=nmodel]{ensemble_lm.dat};
\addlegendentry{Heuristic}
\addlegendentry{Both}
\end{axis}
\end{tikzpicture}
\end{minipage}
\begin{minipage}{.49\linewidth}
    \centering
  \begin{tikzpicture}
\begin{axis}[
  title={\small({\bf b}) Dev other},
  xlabel={\small Number of Models},
  ylabel={\small WER},
  xtick=data,
  width=1.1\columnwidth,
  height=4cm,
  ticklabel style={font=\scriptsize},
  legend style={font=\scriptsize}
]
\addplot table [y=other_heuristic, x=nmodel]{ensemble_lm.dat};
\addplot table [y=other_filter, x=nmodel]{ensemble_lm.dat};
\addlegendentry{Heuristic}
\addlegendentry{Both}
\end{axis}
\end{tikzpicture}
\end{minipage}
\caption{WER with respect to number of models in ensemble under the clean ((\textbf{a})) or noisy ((\textbf{b})) setting. Results are with LM beam search decoding and averaged across three runs. (\textit{Both:} heuristic and confidence-based filters)}
\label{fig:ensembles}
\vspace{-4mm}
\end{figure}


\begin{table*}
    \setlength{\tabcolsep}{5.6pt}
    \centering
    \begin{tabular}{l c c c c | c c c c} 
        \toprule
         \multirow{3}{*}{Method} & \multicolumn{4}{c|}{No LM} & \multicolumn{4}{c}{With LM} \\
          & \multicolumn{2}{c}{Dev WER} & \multicolumn{2}{c|}{Test WER (WRR)} & \multicolumn{2}{c}{Dev WER} & \multicolumn{2}{c}{Test WER (WRR)} \\
         & clean & other & clean & other & clean & other & clean & other \\
        \midrule
        Baseline Paired 100hr & 14.00 & 37.02 & 14.85 & 39.95 & 7.78 & 28.15 & 8.06 & 30.44 \\
        \midrule
        \multicolumn{9}{c}{Paired 100hr + Unpaired 360hr clean speech} \\
        \midrule
        Oracle & 7.20 & 25.32 & 7.99 & 26.59 & 3.98 & 17.00 & 4.23 & 17.36 \\ 
        Single Pseudo & 9.61 & 29.72 & 10.27 (66.8\%) & 30.50 (70.7\%) & 5.84 & 21.86 & 6.46 (41.8\%)  & 22.90 (57.6\%)  \\
        Ensemble (5 models) & 9.00 & 27.74 & \textbf{9.62} (76.2\%)  & 29.53 (78.0\%) & 5.41 & 20.31  & \textbf{5.79} (59.3\%) & 21.63 (67.4\%) \\
        \midrule
        \multicolumn{9}{c}{Paired 100hr + Unpaired 500hr noisy speech} \\
        \midrule
        Oracle & 6.90 & 17.55 & 7.09 & 18.36 & 3.74 & 10.49 & 3.83 & 11.28 \\
        Single Pseudo & 10.90 & 28.37 & 11.48 (43.4\%) & 29.73 (47.3\%) & 6.38 & 19.98 & 6.56 (35.5\%)  & 22.09 (43.6\%) \\
        Ensemble (4 models) & 10.41 & 27.00 & 10.50 (56.1\%)  & \textbf{29.25} (49.6\%)  & 6.01 & 18.95 & 6.20 (44.0\%) & \textbf{20.11} (53.9\%)  \\
        \bottomrule
    \end{tabular}
    \caption{Best results from single runs tuned on the dev sets. The best filtering setup found in Section~\ref{sec:filtering-result} is applied.}
    \label{tab:wers}
\vspace{-2.5mm}
\end{table*}


\begin{table}
    \setlength{\tabcolsep}{4pt}
    \centering
    \begin{tabular}{l c c c} 
        \toprule
        & & No LM & With LM \\
        \multirow{2}{*}{Method} & Text & Test clean & Test clean \\
         & (\# words) & WER (WRR) & WER (WRR)\\
        \midrule
        Cycle TTE~\cite{hori2019cycle} & 4.8M & 21.5 (27.6\%) & 19.5 (30.6\%$^\ast$) \\
        ASR+TTS~\cite{baskar2019self} & 3.6M & 17.5 (38.0\%) & 16.6 (-) \\
        this work & 842.5M & \textbf{9.62} (\textbf{76.2\%}) & \textbf{5.79} (\textbf{59.3\%}) \\
        \bottomrule
    \end{tabular}
    \caption{A comparison with previous work using 100hr paired data and 360hr unpaired audio. WRR is computed with the baseline and oracle WER from the original work if available. ($^\ast$: The oracle WER is without LM decoding, so the WRR is an upper bound estimation.) }
    \label{tab:wers-comparison}
\vspace{-3mm}
\end{table}

\subsubsection{Importance of Filtering}
\label{sec:filtering-result}

Figure~\ref{fig:filtering} shows various filtering functions and the resulting amount of data, the quality of the labels and the corresponding model performance. Label quality is defined as the WER of the filtered pseudo-labels as compared to the ground truth. We apply our heuristic filtering,~i.e.~``no \textsc{EOS} + $n$-gram'' filters, with $c\!=\!2$ and $n\!=\!4$  and then add confidence-based filtering on top of the filtered data set. We can see that filtering indeed improves the pseudo-label quality as we adjust the threshold on the confidence score.

In the clean setting, the heuristic filter removes 1.8\% of the data, and further removal of the worst 10\% of the pseudo-labels based on confidence scores results in a 5.2\% relative improvement in WER on the dev clean set compared with a baseline without filtering. More aggressive filtering improves the label quality but results in worse model performance.

In the noisy setting, removing the worst 10\% of the pseudo-labels results in a significant reduction in WER, and the best performance comes from filtering 60\% of the labels with a WER 22.7\% relative lower on the dev other set compared with no filtering. Filtering more data leads to the same degradation in model performance as in the clean setting.

\subsubsection{Model Ensembles}

Figure~\ref{fig:ensembles} shows WER as a function of the number of models in the ensemble on the dev sets for both clean and noisy settings. We can see that combining multiple models improves the performance, especially for the noisy setting, where we obtain a 13.7\% relative improvement with six models and heuristic filtering. One possible explanation is that since the \textit{sample ensemble} uses different transcripts for the same utterance at training time, this keeps the model from being overly confident in a noisy pseudo-label. We also show that the two filtering techniques can be combined with ensembles effectively. In the noisy setting, model ensembles with both filterings improve WER by 27.0\% relative compared with a single model without any filtering (Figure~\ref{fig:filtering}(d)).

\subsubsection{Comparison with Literature}

Table~\ref{tab:wers} summarizes our best results, as well as the supervised baseline and the oracle models trained with ground-truth transcriptions. We present results from both AM only greedy decoding and LM beam search decoding to demonstrate the full potential of self-training. In addition to WER, we report WER recovery rate (WRR)~\cite{ma2008unsupervised} to demonstrate how much gap between the baseline and the oracle that we can bridge with pseudo-labels. WRR is defined as
\begin{equation*}
     \frac{\textrm{baseline WER} - \textrm{semi-supervised WER}}{\textrm{baseline WER} - \textrm{oracle WER}}.
\end{equation*}
When decoded with an external LM, our best model achieves a WRR over 50\% in both clean and noisy speech settings.

Table~\ref{tab:wers-comparison} compares our approach with other semi-supervised learning methods with sequence-to-sequence models that use the same audio data setup. We see that our conventional pseudo-labelling approach together with filtering and ensemble produces a WER at least 65.1\% relatively lower than the previously best results. The gain comes from the strong baseline model with TDS-based encoders~\cite{hannun2019sequence} to generate the pseudo-labels, and a much larger unpaired text corpus, which we believe is easy to obtain in a real-world setting. As a comparison, the baseline WER on the test clean set is above 20 in~\cite{hori2019cycle,baskar2019self}. However, even with a strong baseline, we achieve a WRR at least 93.8\% relatively higher than other methods.

\vspace{-2mm}
\section{Related Work}
\vspace{-3mm}


In speech recognition, self-training has been explored in hybrid systems~\cite{charlet2001confidence,wessel2004unsupervised,vesely2013semi,vesely2017semi,kemp1999unsupervised}. Prior work mainly focuses on different ways of data filtering to improve pseudo-label quality, e.g.~confidence-based filtering~\cite{charlet2001confidence,wessel2004unsupervised} and agreement-based selection~\cite{de2016high}, which also takes advantage of multiple systems. The data selection process can take place at different levels ranging from frames to utterances~\cite{vesely2013semi,vesely2017semi}. In~\cite{parthasarathi2019lessons}, the output probability of a teacher model is used as soft pseudo-labels to train a student model. Training with pseudo-labels can give an improvement to WER not only for low-resource languages~\cite{vesely2013semi,vesely2017semi} but also on large-scale data sets~\cite{parthasarathi2019lessons}.

Recently-proposed semi-supervised approaches for end-to-end speech recognition take advantage of text-to-speech (TTS) modules to generate synthetic data from unpaired text~\cite{hayashi2018back} or introduce a cycle-consistency loss between the input and the output of an ASR+TTS pipeline~\cite{hori2019cycle,baskar2019self}. Alternatively, inter-domain loss is proposed to constrain speech and text in the same embedding space~\cite{karita2018semi}. In this work, we demonstrate that the self-training approach is simple yet effective with end-to-end systems.


\vspace{-3mm}
\section{Conclusion}
\vspace{-3mm}

We have shown that self-training can yield substantial improvements for end-to-end systems over a strong baseline model by leveraging a large unlabelled data set. We show that filtering mechanisms tailored to the types of mistakes encountered with sequence-to-sequence models as well as an ensemble of models can further improve the accuracy gains from self-training. Our experiments on LibriSpeech have set forth a strong baseline model and a reproducible semi-supervised learning setting for which new and more sophisticated approaches can be evaluated.




\vspace{-2mm}
\section{Acknowledgements}
\vspace{-2mm}
Thanks to Tatiana Likhomanenko, Qiantong Xu, Ronan Collobert and Gabriel Synnaeve for their help with this work.

\newpage
\appendix
\section{Appendix}
\label{sec:appendix}
\subsection{Supervised Baseline}
\label{sec:appendix_baseline}

\begin{table}
    \centering
    \begin{tabular}{l c c c c} 
        \toprule
          & \multicolumn{2}{c}{Dev WER} & \multicolumn{2}{c}{Test WER} \\
          & clean & other & clean & other \\
        \midrule
        Liu et al.~\cite{liu2019adversarial} & 21.6 & - & 21.7 & - \\
        Hayashi et al.~\cite{hayashi2018back} & 24.9 & - & 25.2 & - \\
        L{\"u}scher et al.~\cite{luscher2019rwth} & 14.7 & 38.5 & 14.7 & 40.8 \\
        \midrule
        Our model & 14.0 & 37.0 & 14.9 & 40.0 \\
        \bottomrule
    \end{tabular}
    \caption{The WER for various end-to-end models trained on the ``train-clean-100" subset of LibriSpeech. All numbers are reported without an external LM.}
    \label{tab:baseline}
\end{table}
A common setup for semi-supervised ASR is to use the ``train-clean-100" subset of LibriSpeech as the labelled data set~\cite{liu2019adversarial,hayashi2018back}. Table~\ref{tab:baseline} shows the WER from our supervised baseline on ``train-clean-100" as well as several other results from the literature. Hayashi et~al.~\cite{hayashi2018back} use a sequence-to-sequence model with a BiLSTM-based encoder and location-based attention. They train their model on ``train-clean-100" as the baseline for a back-translation style approach. Liu et~al.~\cite{liu2019adversarial} augment a sequence-to-sequence model with the CTC loss. Compared with the two, our baseline WER on the clean dev and test sets are lower by more than 30\% relative.

On the other hand, L{\"u}scher et~al.~\cite{luscher2019rwth} use the sequence-to-sequence model proposed in~\cite{zeyer2018} and to our knowledge, produce the best prior result when limited to ``train-clean-100." Compared with this, our TDS baseline model achieves better WER on the dev sets and has similar test WER. We believe our supervised baseline is a challenging yet practical starting point for semi-supervised experiments. This baseline enables us to more meaningfully demonstrate the improvement from adding additional unlabelled audio or text data.

\subsection{Evaluating Beam Search}
To study the importance of the stable beam search, we evaluate self-training in two other conditions. First, we compare to pseudo-labels generated from the greedy output of the acoustic model alone. We perform greedy decoding with the supervised baseline model on ``train-clean-360" to generate the pseudo-labels. Second, we compare to pseudo-labels generated from a simple beam search with a language model but without the \textsc{EOS} threshold and hard attention limit described in Section~\ref{sec:inference}.

With each setting, we train three models and report the average WER without an external LM in Table~\ref{tab:beam}. We also compare the pseudo-labels with the ground-truth transcription of ``train-clean-360" and compute label WER as an indicator of the label quality. We can see in Table~\ref{tab:beam} that using an LM in a simple beam search improves the quality of the pseudo-labels and hence the resulting trained model. The stable beam search further improves the pseudo-label quality and the resulting WER of a model trained with those labels.

\begin{table}
    \centering
    \begin{tabular}{l c c c c} 
        \toprule
         Labelling   & Label & Dev Clean & Dev Other \\
         Method & WER    & WER       &  WER \\
        \midrule
        AM greedy & 14.45 & 12.27 & 33.42 \\
        AM+LM simple & 12.15 & 9.73 & 29.77 \\
        AM+LM stable & 8.25 & 9.55 & 28.91 \\
        \bottomrule
    \end{tabular}
    \caption{We show the WER from training on pseudo-labels generated using three approaches: (1) only using the AM without a beam search (AM greedy), (2) using both the AM and the LM with a simple beam search (AM+LM simple) and (3) using both the AM and LM with the stable beam search described in Section~\ref{sec:inference} (AM+LM stable). Dev clean and other WERs are reported without an LM averaged over three models.}
    \label{tab:beam}
\end{table}

\subsection{Importance of the LM}
\label{sec:appendix_lmimportance}

\begin{figure}
\begin{minipage}{.49\linewidth}
  \begin{tikzpicture}
\begin{axis}[
  title=({\bf a}) dev clean,
  xlabel=Perplexity,
  ylabel=WER,
  ylabel shift=-3 pt,
  ymax=13,
  width=1.2\columnwidth,
  height=4cm,
]
\addplot table [y=devclean, x=perplexity]{lms.dat};
\end{axis}
\end{tikzpicture}
\end{minipage}
\hspace{3.5mm}
\begin{minipage}{.45\linewidth}
  \begin{tikzpicture}
\begin{axis}[
  title=({\bf b}) dev other,
  xlabel=Perplexity,
  ymax=34,
  width=1.3\columnwidth,
  height=4 cm,
]
\addplot table [y=devother, x=perplexity]{lms.dat};
\end{axis}
\end{tikzpicture}
\end{minipage}
\caption{The WER on the clean and other development sets as a function of the perplexity of the LM used to generate the pseudo-labels. For each pseudo-labelled data set we train three models and report the average WER without an LM.}
\label{fig:lms}
\end{figure}
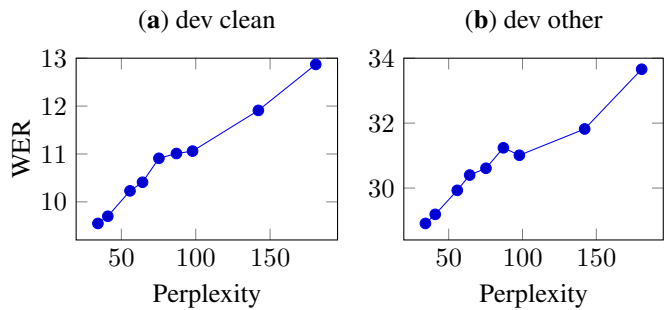

We examine the impact of the LM by training multiple models with pseudo-labels generated from LMs with different perplexity on the dev set. We control for LM perplexity by training the model for a variable number of steps. For each pseudo-label set, we train three models and report the average WER without decoding with an LM. 

In Figure~\ref{fig:lms} we show the reduction in WER from self-training on pseudo-labels generated with varying LM perplexities. We can see a clear trend that when the LM perplexity decreases, the WER on the dev set also decreases. In other words, a better LM leads to better model performance for self-training. In Table~\ref{tab:beam} we show that without using any language model to generate pseudo-labels (AM Greedy), we get a WER of 12.27 on dev clean and 33.42 on dev other. Compared with Figure~\ref{fig:lms}, it is clear that using an LM even with higher perplexity improves the effectiveness of self-training. 


\bibliographystyle{IEEEbib}
\bibliography{refs}

\end{document}